\documentclass[journal]{IEEEtai}

\usepackage[colorlinks,urlcolor=blue,linkcolor=blue,citecolor=blue]{hyperref}

\usepackage{color,array}

\usepackage{graphicx}

\usepackage{algorithm}
\usepackage{algorithmic}

\usepackage{amsmath, amssymb}
\usepackage{multirow}
\usepackage{booktabs}
\usepackage{arydshln}
\usepackage{tabularx}

\DeclareMathOperator*{\argmax}{argmax}

\begin{document}

\title{EAUWSeg: Eliminating annotation uncertainty in weakly-supervised medical image segmentation}

\author{Lituan Wang, Lei Zhang, \IEEEmembership{Senior Member, IEEE}, Yan Wang, Zhenbin Wang, Zhenwei Zhang, and Zhang Yi, \IEEEmembership{Fellow, IEEE}
\thanks{This work was supported by the National Natural Science Foundation of China under Grant 62025601, and Grant 62376174. }
\thanks{Lituan Wang, Lei Zhang, Zhenbin Wang, Zhenwei Zhang, and Zhang Yi are with the Machine Intelligence Laboratory, College of Computer Science, Sichuan University, Chengdu 610065, China. E-mail: lituanwang@scu.edu.cn; leizhang@scu.edu.cn; wangzhenbin@stu.scu.edu.cn; zhangzw@stu.scu.edu.cn; zhangyi@scu.edu.cn. 
Yan Wang is with the Institute of High Performance Computing, A*STAR, Singapore 138632. E-mail: wangyan@ihpc.a-star.edu.sg. (Corresponding author: Lei Zhang).}}


\maketitle

\begin{abstract}
Weakly-supervised medical image segmentation is gaining traction as it requires only rough annotations rather than accurate pixel-to-pixel labels, thereby reducing the workload for specialists. Although some progress has been made, there is still a considerable performance gap between the label-efficient methods and fully-supervised one, which can be attributed to the uncertainty nature of these weak labels. To address this issue, we propose a novel weak annotation method coupled with its learning framework EAUWSeg to eliminate the annotation uncertainty. Specifically, we first propose the Bounded Polygon Annotation (BPAnno) by simply labeling two polygons for a lesion. Then, the tailored learning mechanism that explicitly treat bounded polygons as two separated annotations is proposed to learn invariant feature by providing adversarial supervision signal for model training. Subsequently, a confidence-auxiliary consistency learner incorporates with a classification-guided confidence generator is designed to provide reliable supervision signal for pixels in uncertain region by leveraging the feature presentation consistency across pixels within the same category as well as class-specific information encapsulated in bounded polygons annotation. 
Experimental results demonstrate that EAUWSeg outperforms existing weakly-supervised segmentation methods. Furthermore, compared to fully-supervised counterparts, the proposed method not only delivers superior performance but also costs much less annotation workload. This underscores the superiority and effectiveness of our approach.

\end{abstract}

\begin{IEEEImpStatement}
Benefit to its ability of reducing annotation workload, label-efficient methods have gaining traction in weakly-supervised medical image segmentation. We revisit existing label-efficient medical image segmentation methods and observe that these weak labels introduce considerable uncertainty for segmentation model constructing, which leads to considerable performance gap between the label-efficient methods and fully-supervised one. To address this problem, a novel weak annotation method BPAnno that simply labeling two polygons for a lesion, and its coupled learning framework EAUWSeg is proposed to eliminate the annotation uncertainty. Extensive experiments demonstrate that our EAUWSeg can achieve superior performance while with less than 20\% of the annotation workload when compared to fully-supervised counterparts. This reveals that the proposed method can be a cost-effective solution for improving the performance in weakly-supervised medical image segmentation.
\end{IEEEImpStatement}


\section{Introduction}

\IEEEPARstart{M}{edical} image segmentation plays a crucial role in biomedical image analysis \cite{han2024dmsps}, such as diagnosis, treatment, and radiotherapy planning. As manual segmentation is usually labor-intensive, time-consuming and rely on professional domain knowledge \cite{zhai2023pa}, automatic medical image segmentation has been widely dedicated and series methods have been proposed. However, the successes of existing methods rely mainly on large-scale meticulously annotated data, which requires significant domain expertise as well as expensive annotation cost. 

\begin{figure}[!h]
\centering
\includegraphics[width=0.495\textwidth]{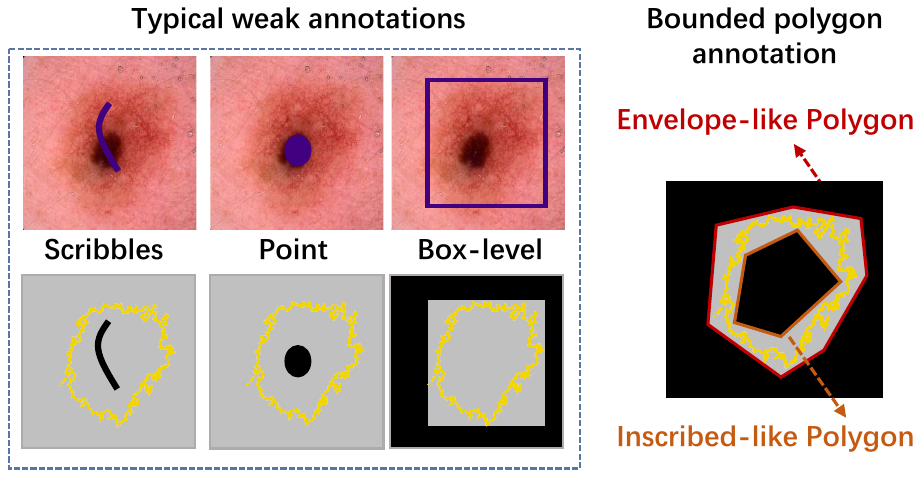}
\caption{Comparison of the typical weak annotation methods and our proposed bounded polygon annotations, including the annotation strategies and the annotation uncertainty. The yellow curves show groundtruth segmentation. The black and gray denote the certain and uncertain regions, respectively. }
\label{fig:fig1}
\end{figure}

To alleviate the burdens associated with image annotation, weakly-supervised medical image segmentation is gaining traction as it requires only weak or sparse annotations \cite{gao2022segmentation}, such as image-level labels \cite{wu2019weakly}, scribbles \cite{lin2016scribblesup}, bounding boxes \cite{rajchl2016deepcut}, and point annotations \cite{dorent2021inter}. 
Although some progress has been made by using label-efficient annotations for training, there is still a considerable performance gap between the label-efficient methods and fully-supervised ones \cite{li2023scribblevc}.
We revisit existing label-efficient medical image segmentation methods and observe that these weak labels introduce considerable uncertainty for segmentation model constructing. Fig. \ref{fig:fig1} provides the visual representation of the supervision signals introduced by different label-efficient annotations, in which most information (defined by the gray region) are uncertain. 
The uncertainty supervision signals provided by label-efficient annotations may induce model training oscillations, thus impair the training of the model to approach the performance achieved in a fully supervised manner \cite{wang2023blpseg}. 
Consequently, there is an urgent need to explore label-efficient methods that can reduce annotation uncertainty, and develop methods that can assistant to eliminate the label uncertainty during model training.



In this work, we propose a novel weak annotation method coupled with its learning framework to eliminate the annotation uncertainty, and facilitate stable training in the weakly-supervised medical image segmentation with more reliable supervision signal. To this end, we introduce the bounded polygons annotation, which simply requires labeling two polygons that are similar to the inscribed and outer envelope-like delineations of lesion (as shown in Fig. \ref{fig:fig1}). The proposed bounded polygons annotation has three advantages: (1) it reduces the label burden compared with pixel-to-pixel accurate labels, (2) it restricts the uncertainty information to gray region between two polygons, (3) it explicitly provides prior emphasis on lesion boundaries during model training. 
Tailored for the proposed weak annotation, we propose a EAUWSeg method to further eliminate the uncertainty included in the bounded polygon annotation by explicitly treating bounded polygons as two separated annotations. For the envelope-like annotation, pixels within red contour belong to foreground class, otherwise belong to background class. For the inscribed-like annotation, pixels within purple contour belong to foreground class, otherwise belong to background class. In this way, the uncertainty region provides adversarial supervision signal for model training to learn invariant feature. Then, by leveraging the existing observation that similar pixels in the feature space prefer to generate consistent category predictions \cite{wu2023sparsely}, we design a Classification-guided Confidence Generator (CCG) to measure the feature similarity between certain and uncertain pixels from a probabilistic perspective. 
Moreover, we adopt a Confidence-auxiliary Consistency Learner (CCL) that prefers to ensure the accuracy of certain pixels and attract uncertain pixels with the same category to preserve consistency feature representation. In the collaboration of CCG and CCL, more reliable supervision signal in uncertain region can be provided during model training to facilitate stable training in the weakly-supervised medical image segmentation.

Overall, our contributions can be summarized as follows: 
\begin{enumerate}
  \item We propose a novel weak annotation method that labels only two bounded polygons and the coupled learning framework for medical image segmentation, which further eliminate the annotation uncertainty existed in most label-efficient methods.
  \item We propose the tailored learning mechanism that explicitly treat bounded polygons as two separated annotations, which can provide adversarial supervision signal for model training to learn invariant feature.
  \item We propose a Confidence-auxiliary Consistency Learner that incorporates with a Classification-guided Confidence Generator to provide reliable supervision signal for pixels in uncertain region by leveraging the intra-class similarity and inter-class discriminative from both the feature and category perspective. It is worth noting that the CCL and CCG modules will be discarded during inference, which not increase computation complexity. 
  \item To evaluate our method, we provide the bounded polygon annotations on two widely used medial image segmentation datasets, i.e., ISIC2017 \cite{codella2018skin} and Kvasir-SEG \cite{jha2020kvasir}. Extensive experiments on these two datasets demonstrate that our EAUWSeg outperforms existing weakly-supervised segmentation methods. Furthermore, the proposed method delivers superior performance with less than 20\% of the annotation workload when compared to fully-supervised counterparts. These results reveal that bounded polygon annotations coupled with EAUWSeg can be a cost-effective solution for the segmentation performance preserving.
\end{enumerate}


\section{Related Works}
\subsection{Weakly-supervised Medical Image Segmentation}
Without the requirement of large densely annotated data, weakly-supervised learning has gained significant attention in medical image segmentation \cite{tajbakhsh2020embracing, zhai2023pa}. As the most efficient weak annotation method, image-level annotations only require classification labels and generates class activation maps \cite{zhou2016learning} for training. Although image-level annotations method is convenient, it has limited performance due to the extremely weak supervision \cite{wang2023intra}. Box-level annotation is usually defined by two corner coordinates, which provides localization-awareness compared to the image-level annotation \cite{hsu2019weakly}. However, boxes for different objects may tend to overlap with each other, making it difficult to accurately approximate the target boundary, especially for complex shapes \cite{wu2023sparsely}. Point annotations provide a small number of pixels for different classes and can better handle complex shapes, which may be more preferable to medical segmentation compared to box-level annotations. Despite its efficiency, the segmentation model trained with point annotations tends to overfit the small number of annotated pixels when comparing with the large number of unannotated pixels.

Scribble-based annotations provide labels for a sparse set of pixels of each class for training, and are usually more obtainable in medical image segmentation by considering its annotation efficiency, performance effectiveness as well as the friendliness to the annotation of nested structure \cite{valvano2021learning}. 
Only the scribbles of the background and each object are given, while the groundtruth of other pixels remains unknown, which is harmful to the segmentation performance. An intuitive resolution is to expand the scribble annotations by considering the prior assumptions \cite{ji2019scribble} or using the learned foreground features through the deep neural networks \cite{zhang2022cyclemix}. However, due to the lack of supervisory signals, the constructed models usually fail to capture the object structure and confuse on the object boundary. 
To address this issue, a series of studies have concentrated on learning adversarial shape priors at the expense of requiring additional fully-annotated masks \cite{valvano2021learning}. However, acquiring such fully annotated datasets may present challenges in many clinical practices, rendering these existing methods both  costly and lacking in scalability. 
Our work aims to explore new weak annotation method that can prompt the performance of automated medical image segmentation without auxiliary datasets.  

\subsection{Contrastive Learning}
Contrastive learning argues that similar samples should have similar representations, and the representations of different samples should be different \cite{chen2020simclr}. Based on this, contrastive loss is usually designed to enforce representations to be similar for similar pairs and dissimilar for dissimilar pairs \cite{jaiswal2020survey}. Considering its powerful self-supervised feature extracting ability from the unlabeled data, contrastive learning has been widely used in many image-level tasks. Among all these methods, the key is the selection mechanism designing of contrastive sample pairs, i.e., positive and negative pairs.

Recently, contrastive learning has been extended from image-level task to pixel-level ones to mine informative information from unlabeled data \cite{NEURIPS2023_1f7e6d5c, wang2021exploring}. As mentioned earlier, constructing contrastive sample pairs is crucial for discriminative feature learning.  In the context of pixel-level tasks, sample pairs are usually constructed through pseudo labels or spatial structure, which may introduce noisy sampling. To alleviate this problem, prediction uncertainty has been injected into the sampling to reduce the number of noisy samples \cite{wang2022uncertainty}.

\begin{figure*}[htb]
\centering
\includegraphics[width=0.9\textwidth]{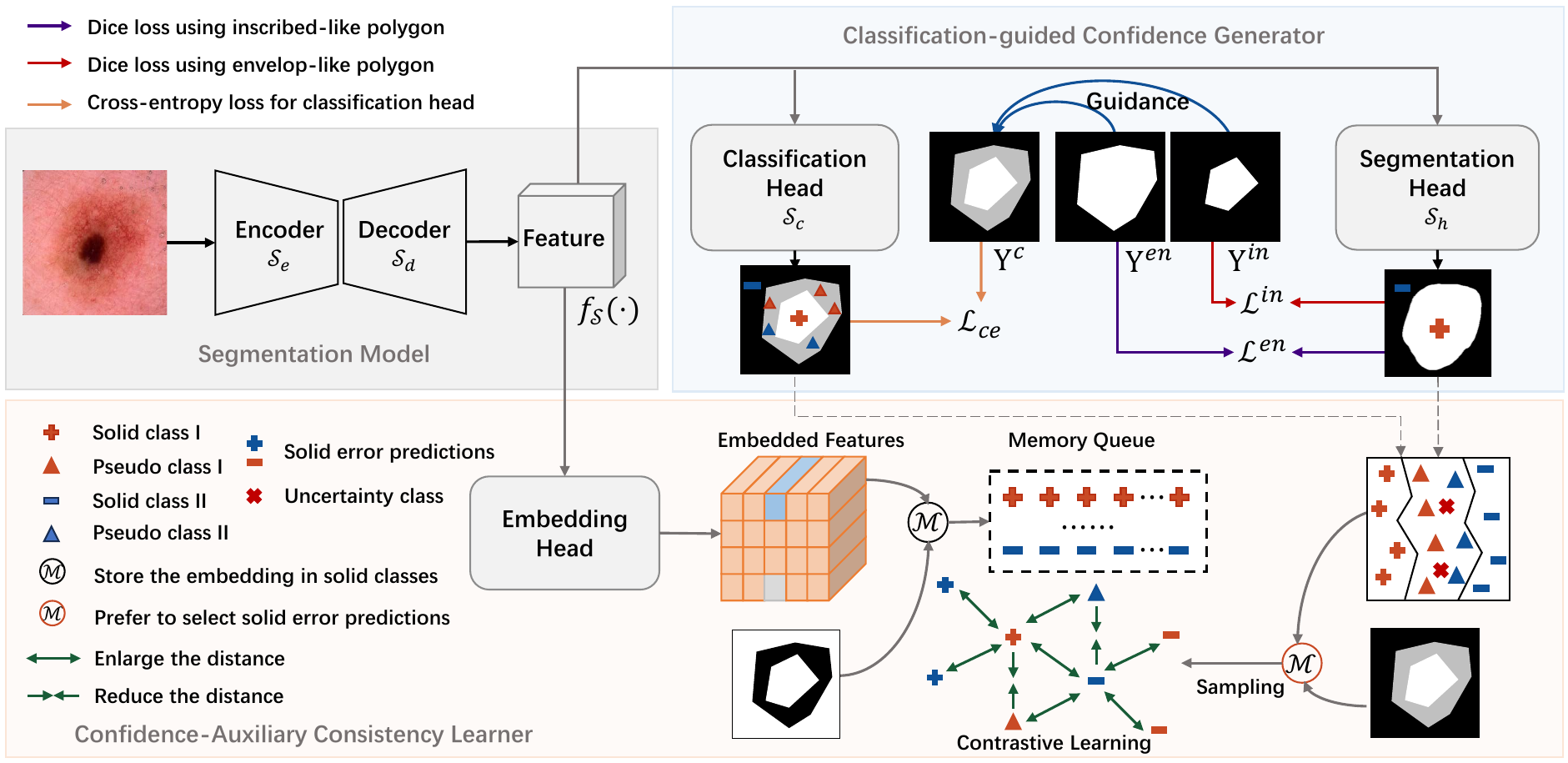}
\caption{The overall framework of our proposed EAUWSeg. It includes a segmentation model supervised by two bounded polygons and a multi-class classification labels. Additionally, a confidence-auxiliary consistency learner is integrated to focus on compact feature learning in the uncertain region. During training, the extracted feature $f_{\mathcal{S}}(\cdot)$ is input into the segmentation head to generate feature representation for lesions. Simultaneously, the classification head supervised by $y^{c}$ is used to generate the confidence of uncertain pixels, the embedding head guided by bounded polygon annotations and confidence of uncertain pixels is utilized to construct a compact feature space. }
\label{fig:fig2}
\end{figure*}

\section{Method}
In this work, we propose a novel bounded polygon annotation method, i.e., \textit{BPAnno}, and its corresponding segmentation framework, i.e., \textit{EAUWSeg}, to eliminate annotation uncertainty in weakly-supervised medical image segmentation. Our EAUWSeg is in general applicable for many existing medical image segmentation models, such as UNet \cite{ronneberger2015u}, DeepLabV3+ \cite{chen2018encoder}, TransUNet \cite{CHEN2024103280}, with encoder and decoder phases. The overall framework is illustrated in Fig. \ref{fig:fig2}. 

\subsection{Problem Setting and Bounded Polygons Annotation}
\label{sec:3.1}

In the scenario of classical weakly-supervised segmentation, the input pixels $x$ are usually divided into the labeled pixels $x_{l}$ and unlabeled pixels $x_{ul}$. In this way, the corresponding labels $y_{l}$ for the labeled pixels $x_{l}$ will be directly used for supervision by employing the partial cross-entropy loss, which can be formulated as follows: 
\begin{equation}
\label{eqn:slloss}
\mathcal{L}_{l}(p, y) = -\displaystyle{\sum\limits_{y \in y_{l}}}ylog(p),
\end{equation}
where $p$ is the segmentation prediction. For the unlabeled pixels, there is no off-the-shelf label for supervision, and a lot of work focus on assigning pseudo labels to unlabeled pixels for supervision \cite{wu2023exploring, liu2023contrastive}.  
The overall objective function can be formulated as follow:
\begin{equation}
\label{eqn:wslloss}
\mathcal{L} = \mathcal{L}_{l} + \mathcal{L}_{ul}.
\end{equation}
However, assigning pseudo labels to unlabeled pixels not only requires a time-consuming multi-stage training process, but also results in misleading or biases \cite{wu2023sparsely}.

To address this problem, this work introduces the bounded polygon annotation method that simply requires labeling two polygons that are similar to the inscribed and outer envelope-like delineations of lesion (as shown in Fig. \ref{fig:fig1}). To further eliminate the uncertainty included in the bounded polygon annotation, we explicitly treat bounded polygons as two separated annotation, i.e., inscribed-like annotation $y^{in}$ and  envelope-like annotation $y^{en}$. 
Different from the classical weakly annotation methods, the input pixels $x$ are divided into the certain labeled pixels and the uncertain pixels in our bounded polygon annotations. This work aims at providing more reliable supervision signal for pixels in uncertain region during model training.

For convenience, we define $\Omega_{I}$, $\Omega_{\Delta}$, $\Omega_{O}$ as the spatial domain inside the inscribed-like annotation, between the inscribed-like and envelope-like delineations, and outside the envelope-like annotation, respectively. Here, the certain labeled pixels and uncertain pixels can be depicted as $x_{i} \in \Omega_{I} \cup \Omega_{O}$ and $x_{uc} \in \Omega_{\Delta}$, the corresponding labels $y_{i}=1$ if $x_{i} \in \Omega_I$ and $y_{i}=0$ if $x_{i} \in \Omega_O$, otherwise, $y_i$ is uncertain.
The spatial domain of input image $x$ and the envelope-like annotation can be depicted as $\Omega = \Omega_{I} \cup \Omega_{O} \cup \Omega_{\Delta}$ and  $\Omega_{E} = \Omega_{I} \cup \Omega_{\Delta}$, respectively.
In this way, our proposed EAUWSeg tries to learn from the ``certain/uncertain'' pixels instead of ``labeled/unlabeled'' pixels in the classical weakly-supervised segmentation. The objective function in our EAUWSeg can be re-formulated as:
\begin{equation}
\label{eqn:bwslloss}
\mathcal{L} = \mathcal{L}_{c} + \mathcal{L}_{uc}.
\end{equation}
The feature learning of certain pixels have been well solved. Hence, this work focuses on eliminating annotation uncertainty and thus providing reliable supervision signals for pixels in the uncertain regions.


\subsection{Framework of EAUWSeg}
EAUWSeg is tailored for the proposed bounded polygon annotation, and mainly focuses on eliminating annotation uncertainty for pixels belong to $\Omega_\Delta$. As illustrated in Fig. \ref{fig:fig2}, EAUWSeg consists of 1) a mainstream segmentation network supervised by two bounded polygons segmentation labels to implicitly define the certain region and uncertain region during network training, 2) a classification-guided confidence generator to provide the category-level prediction confidence for pixels $x_{i} \in \Omega_\Delta$ by leveraging a tailored multi-class classification task, 3) a confidence-auxiliary consistency learner to distinguish reliable pixels in uncertain region can assign the corresponding ``certain'' labels. 

Let $\mathcal{S}_e$, $\mathcal{S}_d$, and $\mathcal{S}_h$ denote the encoder, the decoder, and segmentation head used in our proposed framework that are parameterized by $\Theta_e$, $\Theta_d$ and $\Theta_h$, respectively. In the proposed EAUWSeg, the bounded polygon annotation is treated as two separate masks, i.e., inscribed-like and envelope-like masks, and the basic segmentation loss function in EAUWSeg can be formulated as: 
\begin{equation}
\label{eqn:lloss}
\mathcal{L}_{c} = \displaystyle{\sum\limits_{x}\left(\mathcal{L}_{in}(p, y^{in}) + \mathcal{L}_{en}(p, y^{en}) \right)},
\end{equation}
where $p$ is the predicted probability maps for input image $x$. In this work, the following dice loss is employed for both $\mathcal{L}_{in}$ and $\mathcal{L}_{en}$:
\begin{equation}
\label{eqn:diceloss}
\mathcal{L}_{dice} = 1-\displaystyle{\frac{2\times\sum_{i=1}^{H\times W \times D}p_{i}y_{i}}{\sum_{i=1}^{H\times W \times D}(p_{i}^{2}+y_{i}^{2})}},
\end{equation}
where $H\times W \times D$ denotes the input image size, $p_{i}$ and $y_{i}$ denotes the prediction probability and label for pixel $i$, respectively.
However, training with $\mathcal{L}_{c}$ will introduce inconsistency supervision signals, since $y^{in}$ and $y^{en}$ have the following characteristics: $y^{in}_i = 1$ for $x_i \in \Omega_{I}$, $y^{en}_i = 1$ for $x_i \in \Omega_{E}$, and others are $0$. In this way, pixels in uncertain region will have different labels during training, i.e., $y^{in}_i = 0$ while $y^{en}_i = 1$ for $x_i \in \Omega_{\Delta}$.

To mitigate the influence and leverage this adversarial supervision signal to learn invariant feature during model training, this work focuses on assign more reliable labels for pixels in uncertain region by utilizing feature representation of certain pixels.
For uncertain pixels, we want to utilize the potential similarity between pixels in the same category, i.e., $x_i \in \Omega_{I} \cup \Omega_O$ and $x_j \in \Omega_{\Delta}$ to mine informative information. With these definitions, the loss function of BPAnno-supervised segmentation can be formulated as:
\begin{equation}
\label{eqn:generalloss}
\mathcal{L} = \mathcal{L}_{c} + \displaystyle{\mathcal{L}_{uc}(x, y^{in}, y^{en})}.
\end{equation}
Here, $\mathcal{L}_{uc}(x, y^{in}, y^{en})$ denotes the loss function for uncertain pixels. 
 
\subsection{Classification-Guided Confidence Generator}
\label{sec:3.3}
The key point for accurate BPAnno-supervised segmentation is reliable labels assigning for pixels in uncertain region. Different from existing methods that focus on iteratively assigning pseudo label for uncertain pixels, we propose to utilize the intra-class similarity and inter-class discriminative from both the feature and category perspective.

An intuitive idea to approximate the confidence for uncertain pixels $x_{i}$ is the predictive entropy that is calculated according to the following equation:
\begin{equation}
\label{eqn:entropy}
\mathcal{E} = -\displaystyle{\sum\limits_{k}P(y_{{i}_{k}}|x, \Theta_s)\log (P(y_{{i}_{k}}|x, \Theta_s)+\epsilon)}
\end{equation}
where $\Theta_{s}=\{\Theta_{e}, \Theta_{d}, \Theta_{h}\}$ are the parameters of standard segmentation network, $\epsilon$ is a constant to avoid overflow. Similar as previous works, prediction with large entropy is considered as the solid uncertain pixels, which will be dropped during the subsequent learning. For clarity, we define the solid uncertain pixels in uncertain region with category of $-1$: 
\begin{equation}
\label{eqn:uncertainty}
\mathcal{U}^{e}_{i} = \left\{
\begin{aligned}
&-1, &\mathcal{E}_{i}\ge \mu \\
&0, &\text{otherwise} \\
\end{aligned}
\right.
\end{equation}
where $\mu$ is a predefined threshold to mask the uncertain labels, and $\mathcal{U}^{e} \in \mathbb{R}^{C\times H\times W}$ is the estimated uncertainty map with the same size as input image.

To assign more reliable labels for uncertain pixels, we propose to explicitly leverage the potential similarity between certain and uncertain pixels by employing a tailored classification task, which aims at removing as much uncertainty as possible. Let $f_\mathcal{S}(x)$ denote the feature representation generated through the encoder and decoder network, $\mathcal{S}_c$ and $\Theta_{c}$ denote the classification head and its corresponding parameters respectively. Previous work \cite{wu2023sparsely} has shown that similar pixels in the feature space preferable to generate consistent category prediction. Based on this, the constructed classification head is used to model a multi-class classification task with the objective function of:
\begin{equation}
\label{eqn:celoss}
\mathcal{L}_{ce} = -\frac{1}{N}\displaystyle{\sum_{i=1}^{N}}y_{i}^{c}\log P\left(y_{i}^{c}|x, \Theta_e, \Theta_d, \Theta_c\right),
\end{equation}
where $y^{c}$ is the classification labels that $y_{i}^{c}=0$ for $x_i \in \Omega_O$, $y_{i}^{c}=1$ for $x_i \in \Omega_\Delta$, and $y_{i}^{c}=2$ for $x_i \in \Omega_I$, and $N=H\times W \times D$. 

During model training, we assume that ``certain" pixels in uncertain region would prefer to generate prediction of $y_{i}^{c}=0$ for background and $y_{i}^{c}=2$ for foreground. 
In this way, the uncertain map generated by the confidence map can be formulated as:   
\begin{equation}
\label{eqn:uncertaintyC}
\mathcal{U}^{c} = \argmax (P(y=0|f_\mathcal{S}(x), \Theta_{c}), P(y=2|f_\mathcal{S}(x), \Theta_{c})) \odot \mathcal{M}_{u},
\end{equation}where $\odot$ refers to the element-wise multiplication, and $\mathcal{M}_{u}$ is a mask with $x_i=1$ for $x_i\in\Omega_\Delta$, and $x_i=0$ otherwise. 
The final confidence for the uncertain pixels can be formulated as:
\begin{equation}
\label{eqn:uncertainty}
\mathcal{U} = \min(\mathcal{U}^{c} + 2\mathcal{U}^{e}, \mathbf{-1}) \odot \mathcal{M}_{u},
\end{equation} 
Here, $\mathcal{U}$ means that both pixels with large predictive entropy, i.e., $\mathcal{E}_{i}\ge \mu$ and with uncertain classification prediction, i.e., pixels with prediction of $1$ for the multi-class classification task, will be considered as solid uncertain and be assigned with label of $-1$.

\subsection{Confidence-Auxiliary Consistency Learner}
\label{sec:3.4}
Confidence-auxiliary consistency learner aims at generating ``certain" information from uncertain region to facilitate stable training.
An intuitive idea is utilizing contrastive learning to reduce the distance between pixels within same category while enlarging the distance between pixels in different categories. This strategy allows us to conduct the pixel-wise contrastive learning. However, the crucial question is the selection of positive and negative samples, especially for pixels in uncertain region. 
To reduce the influence of uncertain information, we propose to utilize the generated confidence for the uncertain pixels and only the solid certain pixels will be considered during the pixel-wise contrastive learning. In this way, the determined pseudo labels can be obtained as follows:
\begin{equation}
\label{eqn:pseudolabel}
\hat{y} = y \odot (\mathbf{1} - \mathcal{M}_{u}) + \mathcal{U}.
\end{equation}

To provide more reliable supervision signal by using the pixel-wise contrastive learning, we follow two guidelines during sample selection: 1) only feature embedding of pixels in the certain region are stored in this study and further be sampled during the computation of contrastive loss; 2) the anchor sampling in this study focuses on hard samples with error prediction for $x_i \in \Omega_{I} \cup \Omega_{O}$, and samples with higher certainty for $x_{i} \in \Omega_{\Delta}$.
The pixel-wise contrastive loss in this work can be defined as:
\begin{eqnarray}
\label{eqn:pclloss}
\mathcal{L}_{PCL} &=& -\frac{1}{\mathcal{P}}\sum_{i\in \mathcal{P}}\frac{1}{|P\backslash \{i\}|}  \\
&\times& \displaystyle{\sum_{p\in P}}\log \frac{\exp(f_{i}\cdot f_{p}/\tau)}{\exp(f_{i}\cdot f_{p}/\tau) + \frac{1}{|N|}\displaystyle{\sum_{n\in N}}\exp(f_{i}\cdot f_{n}/\tau)}, \nonumber 
\end{eqnarray}  where $\mathcal{P}$ contains the indexes of all ``certain'' pixels in the uncertain region; $P$ and $N$ contains the indexes of positive pixels, i.e., pixels has same class with pixel $i$, and negative pixels, i.e., pixels with different labels to pixel $i$, in the certain region, respectively; $\tau$ is a temperature hyper-parameter.

Considering the semantic representation for deep layers and the effective information for uncertain pixels, feature representation before the segmentation head is embedded into a specific feature space and is employed as the prototype vector in contrastive learning. That is to say, $f_{i}$ denotes the feature embedding of pixels $x_i$ that is calculated according to the following equation:
\begin{equation}
\label{eqn:feature}
f_{i} = f_{\mathcal{S}}(x_{i}, \Theta_{e}, \Theta_{d}).
\end{equation} 

\subsection{Training of EAUWSeg}
To summarize, the overall objective function includes two parts: 1) losses for ``certain'' pixels using fully-supervised segmentation setting, 2) confidence-guided contrastive loss for uncertain pixels. At the early stage of training, segmentation model need to learn the feature representation of lesions with the guidance of supervised loss for ``certain'' pixels, i.e., $\mathcal{L}_{c}$. When the segmentation performance gradually improves, the contrastive loss $\mathcal{L}_{PCL}$ combined with a multi-class classification cross-entropy loss $\mathcal{L}_{ce}$ are added to apply constraints on uncertain pixels in same class to preserve consistency feature representation. Therefore, the overall objective function in this work is formulated as:
\begin{equation}
\label{eqn:totalloss}
\mathcal{L} = \mathcal{L}_{c} + \lambda_{1}\mathcal{L}_{PCL} + \lambda_{2}\mathcal{L}_{ce}, 
\end{equation}
where $\lambda_{1}$ and $\lambda_{2}$ are the parameters to control the contribution of confidence-auxiliary consistency learner and classification-guided confidence generator, respectively.

\section{Experiments}
\label{sec:sec4}


\begin{table*}[]
\centering
\caption{Performance comparison with State-of-the-Arts on ISIC2017 and Kvasir-SEG datasets. Bold and underline denote the best and second best results, respectively.} 
\label{tab:totalresults}
\resizebox{0.999\linewidth}{!}{
\renewcommand{\arraystretch}{1.3}{
\begin{tabular}{cccccccccc}
\toprule
\multirow{2}{*}{Methods} & \multirow{2}{*}{Data} & \multicolumn{4}{c}{ISIC2017} & \multicolumn{4}{c}{Kvasir-SEG} \\ \cline{3-6} \cline{7-10} 
 &  & Dice &Jaccard &Accuracy &Sensitivity &Dice &Jaccard &Accuracy &Sensitivity \\ 
 \midrule
\multicolumn{2}{c}{\textit{Fully-supervised Methods}} &  &   &  &  &  &  & & \\
UNet \cite{ronneberger2015u} & mask & 86.11$\pm$.13 & 77.80$\pm$.16 & 93.83$\pm$.03 & 84.61$\pm$.49 & 89.21$\pm$.24 & 83.50$\pm$.09 & \textbf{96.95$\pm$.05} & 91.40$\pm$.39  \\
UNet++ \cite{zhou2019unet++} & mask & 85.75$\pm$.10 & 77.60$\pm$.18 &\textbf{96.60$\pm$.25} & 84.22$\pm$.21 &89.23$\pm$.14 & 83.43$\pm$.06 & 96.78$\pm$.08 &92.09$\pm$.66\\
DeepLabV3+\cite{chen2018encoder} & mask & 86.15$\pm$.10 & 78.06$\pm$.14 & 93.97$\pm$.03 & 83.79$\pm$.29 & 89.04$\pm$.27 & 82.85$\pm$.22 & 96.87$\pm$.11 & 91.73$\pm$.42 \\
TransUNet \cite{CHEN2024103280} & mask & 86.25$\pm$.13 & \underline{78.21$\pm$.13} & 93.88$\pm$.14 &85.57$\pm$.83 & \underline{89.64$\pm$.13} & \textbf{83.86$\pm$.16} & 96.76$\pm$.10 & 91.74$\pm$.54 \\
TransFuseS \cite{zhang2021transfuse} & mask & 86.09$\pm$.27 & 78.01$\pm$.45 & 93.84$\pm$.16 & 85.08$\pm$.92 & 88.15$\pm$.21 & 82.02$\pm$.18 & 96.48$\pm$.07 &89.94$\pm$.64 \\
HiFormer \cite{heidari2023hiformer} & mask & 86.16$\pm$.22 & 78.02$\pm$.30 & 93.84$\pm$.11 & 85.05$\pm$.76 & 89.18$\pm$.22 & 83.41$\pm$.19 & 96.86$\pm$.04 & 90.40$\pm$.63 \\
\midrule
\multicolumn{2}{c}{\textit{Weakly-supervised Methods}} &  &   &  &  &  &  & & \\
PCE \cite{tang2018normalized} & scribbles &80.94$\pm$.08  &71.19$\pm$.10  &91.64$\pm$.02  &80.85$\pm$.71  &77.21$\pm$.46 &66.46$\pm$.41 &93.83$\pm$.10 &80.92$\pm$1.55 \\
TV \cite{javanmardi2016unsupervised} & scribbles &81.14$\pm$.33 &71.50$\pm$.43 &91.83$\pm$.13 &81.13$\pm$.55 &77.01$\pm$.21  &66.24$\pm$.19  &93.75$\pm$.03 &80.32$\pm$.46  \\
GatedCRF \cite{obukhov2019gated} & scribbles &81.02$\pm$.53  &71.25$\pm$.61 &91.64$\pm$.22 &78.30$\pm$.80  &78.63$\pm$.26  &68.43$\pm$.29  &94.12$\pm$.10 &77.43$\pm$.44 \\
Mumford-Shah \cite{kim2019mumford} & scribbles &76.50$\pm$.78 &65.00$\pm$.97 &90.36$\pm$.30 &72.02$\pm$2.82 &69.61±1.49  &57.25±1.76  &92.13$\pm$.31 &68.12±3.91 \\
USTM \cite{liu2022weakly} & scribbles &80.92$\pm$.10 &71.24$\pm$.12  &91.60$\pm$.13 &79.40$\pm$1.33 &76.65$\pm$.18  &65.95$\pm$.21  &93.72$\pm$.02 &79.30$\pm$.60 \\
ScribbleVC \cite{li2023scribblevc} & scribbles &81.07$\pm$.50  &71.40$\pm$.41 &91.85$\pm$.04  &76.38$\pm$1.00  &77.29$\pm$.39 &66.95$\pm$.37 &93.83$\pm$.18  &76.21$\pm$1.04 \\
DMSPS \cite{han2024dmsps} & scribbles &81.50$\pm$.19 &71.86$\pm$.09 &91.90$\pm$.10 &80.68$\pm$.47 &78.21$\pm$.53  &68.04$\pm$.59  &94.02$\pm$.02  &80.78±2.10 \\
TriMix \cite{zheng2022trimix} & scribbles & 82.03$\pm$.11 &72.65$\pm$.12 &91.76$\pm$.12 &80.39$\pm$.78 &84.23$\pm$.11 &75.83$\pm$.26  &95.44$\pm$.08  &83.46$\pm$.42 \\
\hdashline 
UNet & box &82.17$\pm$.10  &71.34$\pm$.18 &91.62$\pm$.05  &90.71$\pm$.49  &76.68$\pm$.06  &64.42$\pm$.10  &91.82$\pm$.37  &93.85$\pm$.78 \\ 
TransUNet & box &82.71$\pm$.27  &72.17$\pm$.35 &91.98$\pm$.16  &\textbf{91.04$\pm$.64}  &78.61$\pm$.23  &66.69$\pm$.12 &92.82$\pm$.40  &\textbf{92.66$\pm$2.11} \\
\hdashline 
UNet & rectangle &85.38$\pm$.03 &76.52$\pm$.11 &93.38$\pm$.10 &89.35$\pm$1.34  &82.51$\pm$.24  &72.73$\pm$.19  &94.81$\pm$.10  &92.76$\pm$.16 \\ 
TransUNet & rectangle  &85.44$\pm$.21 &76.69$\pm$.19 &93.33$\pm$.02 &\underline{89.79$\pm$2.04} &83.40$\pm$.20  &74.10$\pm$.08  &94.57$\pm$.12  &92.10$\pm$.66 \\
\hdashline 
Ours(UNet) & BPAnno &\underline{86.18$\pm$.05} &78.12$\pm$.04 & 93.83$\pm$.04 & 84.45$\pm$.45 &89.30$\pm$.11 & 83.04$\pm$.15 & 96.88$\pm$.03 &91.98$\pm$.32 \\ 
Ours(TransUNet) & BPAnno & \textbf{86.60$\pm$.17} & \textbf{78.61$\pm$.24} & \underline{93.95$\pm$.12} & 87.55$\pm$1.38 & \textbf{89.88$\pm$.19} & \underline{83.85$\pm$.27} & \underline{96.91$\pm$.08} & \multicolumn{1}{c}{\underline{92.15$\pm$.73}} \\
\bottomrule
\end{tabular}%
}}
\end{table*}

\subsection{Experimental Setup} 
\subsubsection{Datasets}
To evaluate the effectiveness of the proposed method, we conduct the comparative experiments on two widely used medical image segmentation datasets, i.e., ISIC2017\cite{codella2018skin} and Kvasir-SEG \cite{jha2020kvasir} datasets. ISIC2017 is a skin lesion segmentation dataset, on which rich results have been reported in literature for comparisons. It contains 2000, 150, and 600 dermoscopic images in train, valid, and test sets respectively. We follow the official split of train and test set during the experiment. Kvasir-SEG contains 1000 gastrointestinal polyp images and the corresponding groundtruth. we randomly split the dataset into two subsets with 800 and 200 images, respectively. Furthermore, to evaluate the generalization ability of the constructed model, we conduct the cross-training evaluation and apply the model trained on ISIC2017 to test on ISIC2018 dataset for skin lesion segmentation without fine-tuning.
ISIC2018 Dataset \cite{codella2019skin} is a expansion of ISIC2017, and it contains 2594, 100 and 1000 images in train, valid, and test sets respectively. It should be noticed that there is no intersection between the test set of ISIC2018 and the train set of ISIC2017.

\subsubsection{Annotation Generation}
For the bounded polygon annotations, we initially generate approximate bounded polygon through dilation-erosion operations by leveraging available groundtruth masks. Subsequently, a manual refinement process is employed to enhance the accuracy of bounded polygon annotation. In the automatic generation phase, we create coarse envelop-like and inscribed-like polygons by employing dilation and erosion operations on the segmentation masks. Specifically, the dilation operation enlarges the masks, while the erosion operation shrinks them. These modified masks serve as a basis for generating polygons. Douglas-Peucker algorithm \cite{douglas1973algorithms} is then applied to derive approximate contours from the dense masks to make the bounded polygon with the limited number of vertices.

To compare with existing weakly-supervised methods, we also generate the scribble, box and rectangle annotation on these two datasets. Following \cite{wong2023scribbleprompt}, we draw random lines by connecting two end points sampled from $\{(u,v)|y_{uv}=1\}$ to simulate the scribbles. Here, $y\in\{0, 1\}^{H\times W}$ is the given groundtruth binary mask. To obtain the box annotation, we use the object detection method. Similarly, we obtain the rectangle annotation for an image that can be filled to create a rectangular mask by identifying the smallest rectangular area that covers the foreground pixels in the groundtruth mask.

\subsubsection{Implementation Details}
All experiments are conducted using PyTorch and NVIDIA GeForce RTX 3090 GPUs. During training, images are resized to $256 \times 256$ for all backbone networks except for TransFuse and HiFormer, which are set as $192 \times 256$ and $224 \times 224$ respectively. For optimizing, we employ the Adam and AdamW optimizer with an initial learning rate of $1e-4$ for CNN-based and Transformer-based backbone networks, respectively. Typically, we set the maximal number of epochs at 100 for ISIC2017 and 300 for Kvasir-SEG, the batch size at 16, and the hyper-parameters are: $\lambda_1=0.3$, $\lambda_2=0.5$, $\tau = 0.1$, and $\epsilon=1e-6$.  

\subsection{Comparison With State-of-the-Arts}
\begin{figure*}[!h]
\centering
\includegraphics[width=0.99\textwidth]{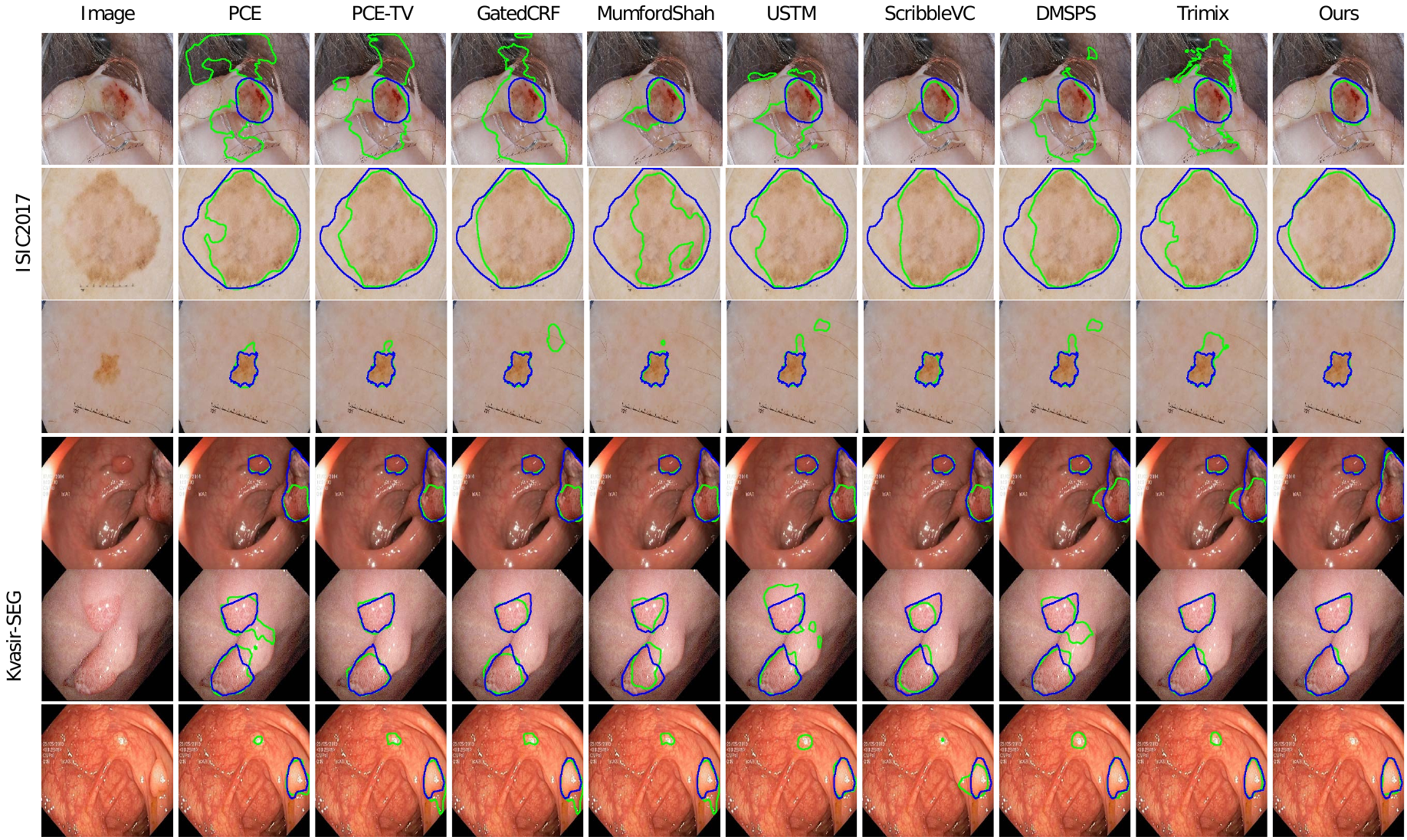}
\caption{Qualitative comparison of different methods on ISIC2017 (top three rows) and Kvasir-SEG (bottom three rows). The green and blue contours indicate the prediction and groundtruth, respectively.}
\label{fig:visual}
\end{figure*}

To demonstrate the comprehensive segmentation performance of our method, we compare EAUWSeg with different state-of-the-art approaches: 
\begin{itemize}
  \item Scribble-supervised methods: 1) different learning strategies on UNet, including partially Cross-Entropy loss \cite{tang2018normalized}, Total Variation loss \cite{javanmardi2016unsupervised}, Gated Conditional Random Field loss \cite{obukhov2019gated}, Mumford-Shah Loss \cite{kim2019mumford}, as well as Uncertainty-aware Self-ensembling and Transformation-consistent Mean Teacher techniques (USTM) \cite{liu2022weakly}. 2) different scribble-supervised frameworks, including ScribbleVC \cite{li2023scribblevc}, DMSPS \cite{han2024dmsps}, and TriMix \cite{zheng2022trimix}.
  \item Box-supervised methods. For fair comparison, we also present the results of classical segmentation networks, i.e., UNet and TransUNet, supervised with bounding box. 
  \item Fully-supervised segmentation methods: 1) CNN-based methods, including UNet \cite{ronneberger2015u}, UNet++ \cite{zhou2019unet++}, and DeepLabV3+ \cite{chen2018encoder}. 2) Transformer-based methods, including TransUNet \cite{CHEN2024103280}, TransFuseS \cite{zhang2021transfuse}, HiFormer \cite{heidari2023hiformer}. Implementation of these networks follow the corresponding github repositories. During training, ResNet50 \cite{he2016deep} is employed as the encoder for UNet and DeepLabV3+, ResNet34 is utilized in the UNet++ and TranFuseS, the default ``R50-ViT-B\_16'' and ``Hiformer-S'' configurations are employed for TransUNet and HiFormer. 
\end{itemize}

Table \ref{tab:totalresults} presents the quantitative evaluation results of the aforementioned methods. For fair comparison with scribble-supervised methods with different learning strategies, we present the results of EAUWSeg with UNet as backbones. 
To demonstrate the effectiveness of EAUWSeg, we also give the results with TransUNet as backbone. The results illustrate that our method outperforms other weakly-supervised methods on both ISIC2017 and Kvasir-Seg datasets, including the scribble-supervised as well as the box-supervised methods.
When compared with the fully-supervised methods, our proposed EAUWSeg can also deliver superior performance, yielding an average Dice score of 86.60\% and 89.88\% on ISIC2017 and Kvasir-Seg, respectively. This underscores the superiority and effectiveness of the proposed BPAnno-supervised strategy and its corresponding learning framework EAUWSeg.
Fig. \ref{fig:visual} shows some qualitative evaluation results, it can be seen that our proposed method achieves better segmentation performance.

\subsection{Ablation Study}

\begin{table}[!h]
\centering
\caption{Comparison of Dice and Sensitivity for six weak annotation methods.} 
\label{tab:ablationboundedAnno}
\resizebox{0.999\linewidth}{!}{
\renewcommand{\arraystretch}{1.3}{
\begin{tabular}{cccccc}
\toprule
\multirow{2}{*}{Data} & \multirow{2}{*}{} & \multicolumn{2}{c}{ISIC2017} & \multicolumn{2}{c}{Kvasir-SEG} \\ \cline{3-6}
 & &Dice &Jaccard &Dice &Jaccard \\ 
\midrule
\multicolumn{2}{c}{\textit{Single Annotation}} &   &  &  & \\  
polygon   &  &85.54$\pm$.20  &76.86$\pm$.09  &85.29$\pm$.29  &76.22$\pm$.51 \\
rectangle &  &85.44$\pm$.21  &76.69$\pm$.19  &83.40$\pm$.20  &74.10$\pm$.08 \\
ellipse   &  &84.61$\pm$.11  &75.35$\pm$.24  &83.61$\pm$.27  &73.80$\pm$.61 \\\midrule
\multicolumn{2}{c}{\textit{Bounded Annotation}}  &   &  &  & \\  
bounded polygon    &  &85.88$\pm$.18 &77.81$\pm$.22  &88.95$\pm$.36  &82.61$\pm$.40 \\
bounded rectangle  &  &85.60$\pm$.22  &77.13$\pm$.28  &88.71$\pm$.10  &82.22$\pm$.27 \\
bounded ellipse    &  &84.63$\pm$.17  &75.56$\pm$.14  &87.78$\pm$.19 &80.76$\pm$.15 \\
\bottomrule
\end{tabular}%
}}
\end{table}

\subsubsection{Effectiveness of the Bounded Annotations}
To analysis the effectiveness of the proposed bounded annotation strategy, we conduct quantitative evaluation of training the TransUNet directly using
different bounded annotation methods, including polygon, rectangle, and ellipse. Considering, box is similar with rectangle, only rectangle is compared since it can achieves better performance. Table \ref{tab:ablationboundedAnno} lists the quantitative comparison based on the Dice score and Sensitivity. The former presents the overall segmentation performance while the latter can reflect the recall of the foreground pixels. 
It can be seen that all these three annotations offer a promising way to initialize the lesion region (with Dice score larger than 80\%), while polygon shows the best performance.
Substituting the single annotation with bounded ones leads to consistent performance improvement for all these annotation methods on two datasets, demonstrating the effectiveness of our proposed bounded-based weak annotation strategy.

\subsubsection{Comparative Analysis of Different Components}
To demonstrate the effectiveness of the proposed component, i.e., confidence-auxiliary consistency learner (CCL) and classification-guided confidence generator (CCG), we carried out the ablation experiments and the results are shown in Table \ref{tab:ablationComponent}. Baseline present the performance of TransUNet trained with bounded polygon annotations. It can be seen that with the gradual introduction of CCL and CCG the performance consistently improves on both ISIC2017 and Kvasir-SEG.

\begin{table}[!h]
\centering
\caption{Ablation Study on ISIC2017 and Kvasir-SEG datasets with TranUNet as the backbone.} 
\label{tab:ablationComponent}
\resizebox{0.999\linewidth}{!}{
\renewcommand{\arraystretch}{1.3}{
\begin{tabular}{ccccccc}
\toprule
\multirow{2}{*}{Baseline} & \multirow{2}{*}{CCL} & \multirow{2}{*}{CCG} & \multicolumn{4}{c}{Evaluation Metrics}  \\ \cline{4-7} 
 & &  &Dice &Jaccard &Accuracy &Sensitivity \\ 
\midrule
\multicolumn{3}{c}{\textit{ISIC2017}} &   &  &  & \\  
\checkmark &  &                      &85.88   &77.81  &93.65  &85.59 \\
\checkmark &\checkmark  &            &86.38   &78.12  &93.81  &86.38 \\
\checkmark &\checkmark  &\checkmark  &86.60   &78.61  &93.95  &87.55 \\\midrule
\multicolumn{3}{c}{\textit{Kvasir-SEG}}  &   &  &  & \\  
\checkmark &  &                      &88.95   &82.61  &96.49  &91.78 \\
\checkmark &\checkmark  &  &89.35  &83.06  &96.90  &92.03 \\
\checkmark &\checkmark  &\checkmark  &89.88   &83.85  &96.91  &92.15 \\
\bottomrule
\end{tabular}%
}}
\end{table}
  
\subsubsection{Comparison With Semi-supervised Methods}

Table \ref{tab:SSL} presents a comparative analysis of our method with five existing semi-supervised segmentation methods on ISIC2017. For these semi-supervised methods, we referred to the results reported in \cite{tang2023consistency}. 
These semi-supervised methods are trained with varying percentages of labeled data ($5\%/10\%/20\%$), and assisted with the rest of unlabeled data ($95\%/90\%/80\%$). While the proposed is trained with only $5\%/10\%/20\%$ samples annotated by bounded polygon, without using the rest of unlabeled data .
Although only supervised with $5\%/10\%/20\%$ samples annotated by bounded polygon, our method outperforms most of the specifically designed SSL methods (except for CASSL) that trained with dense mask and also the rest of unlabeled data, showcasing its robust feature learning capabilities. When compared with CASSL, in which the adversarial training mechanism and the collaborative consistency learning strategy are carefully designed to utilize the unlabeled data, our method has a small performance gap while no need for dense mask and also the unlabeled data. 
This is important to many medical image segmentation tasks since additional unlabeled data may be unavailable in clinical practice.

\begin{table}[!h]
\footnotesize
\centering
\caption{Performance comparison with semi-supervised methods on ISIC2017 test set with Jaccard score as the evaluation metric.}
\label{tab:SSL}
\renewcommand{\arraystretch}{1.3}{
\begin{tabularx}{0.478\textwidth}{ccccc}
\toprule
\multicolumn{1}{c}{Methods}    &\multicolumn{1}{c}{Data}            & 5\% & 10\% & 20\% \\ \midrule
UNet &\multirow{5}{1.5cm}{\centering mask+ \\ unlabeled} & \multirow{1}{0.9cm}{\centering 70.92}  & \multirow{1}{0.9cm}{\centering 71.74}             & \multirow{1}{0.9cm}{\centering 75.27} \\
CLCC \cite{9761710}                        &             & \multirow{1}{0.9cm}{\centering 61.23}  & \multirow{1}{0.9cm}{\centering 65.40}             & \multirow{1}{0.9cm}{\centering 68.93}      \\
MT \cite{tarvainen2017mean}                &             & \multirow{1}{0.9cm}{\centering 73.12}  & \multirow{1}{0.9cm}{\centering74.34}             & \multirow{1}{0.9cm}{\centering 76.98}     \\
ST++ \cite{yang2022st++}                   &                 & \multirow{1}{0.9cm}{\centering 73.26}             & \multirow{1}{0.9cm}{\centering75.51}             & \multirow{1}{0.9cm}{\centering 76.69}     \\
S4-PLCL \cite{alonso2021semi}              &                 & \multirow{1}{0.9cm}{\centering 68.19}             & \multirow{1}{0.9cm}{\centering71.08}             & \multirow{1}{0.9cm}{\centering 71.83}     \\
CASSL \cite{tang2023consistency}           &                 & \multirow{1}{0.9cm}{\centering 76.55}             & \multirow{1}{0.9cm}{\centering77.49}             & \multirow{1}{0.9cm}{\centering 79.31}     \\ 
\midrule
\multirow{1}{2.0cm}{\centering Ours(TransUNet)}   & only BPAnno   & \multirow{1}{0.9cm}{\centering 75.81}    & \multirow{1}{0.9cm}{\centering 76.86}    & \multirow{1}{0.9cm}{\centering 77.54}    \\
\bottomrule
\end{tabularx}}%
\end{table}

\subsubsection{Generalizabilty Analysis With Different Backbones}
\begin{figure}[htb]
\centering
\includegraphics[width=0.495\textwidth]{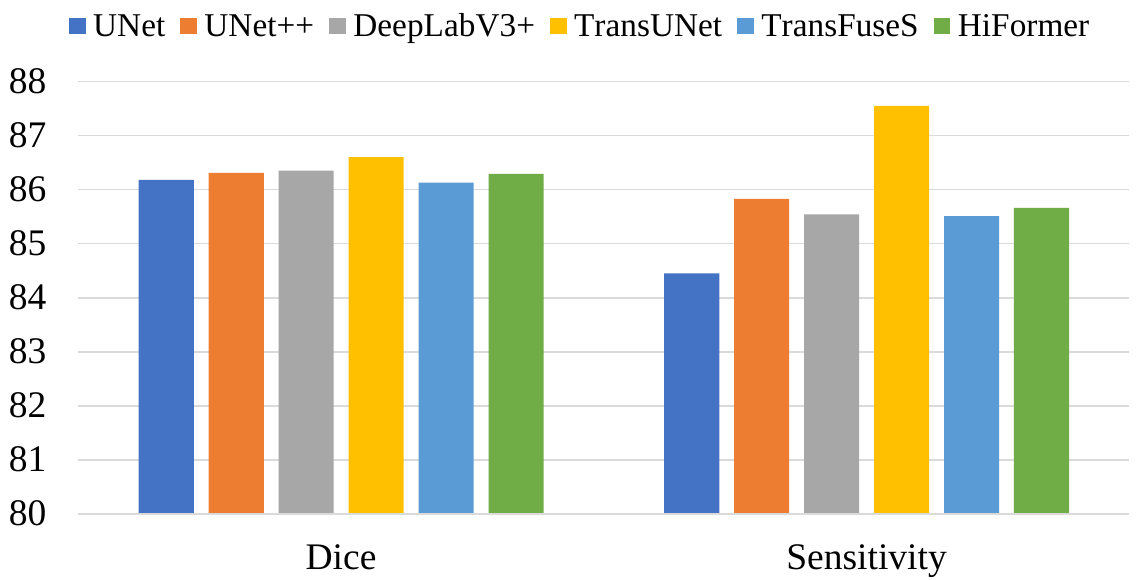}
\caption{Performance comparison of EAUWSeg combined with different backbones on the ISIC2017 test set.}
\label{fig:backbones}
\end{figure}

The proposed EAUWSeg is a plug-and-play model that can be easily combined with different backbones. To demonstrate its generalization ability six widely used segmentation networks are compared, i.e., UNet \cite{ronneberger2015u}, UNet++ \cite{zhou2019unet++}, DeepLabV3+ \cite{chen2018encoder}, TransUNet \cite{CHEN2024103280}, TransFuseS \cite{zhang2021transfuse}, and HiFormer \cite{heidari2023hiformer}. 
From Fig. \ref{fig:backbones}, it can be seen that: 1) the best result is achieved when using TransUNet as the backbone, 2) the proposed method delivers superior performance compared to fully-supervised counterparts as shown in Table \ref{tab:totalresults}. These results reveal that the proposed EAUWSeg generalizes well for different backbones.

\subsubsection{Generalization on ISIC2018}
The generalization ability of the constructed model is important to real application. We evaluate the generalization ability of the proposed method in a cross-training way \cite{yuan2021multi}. Specifically, we apply the model trained on ISIC2017 to test on ISIC2018 dataset for skin lesion segmentation without fine-tuning. 
As presented in Table \ref{tab:generalizabilityISIC2018}, our method achieves comparable generalization performance on ISIC2018 when compared to all the fully-supervised counterparts. This highlights the effectiveness of our EAUWSeg approach as well as the bounded polygons annotation in ensuring robust generalizability.

\begin{table}[!h]
\centering
\caption{Generalizability comparison on ISIC2018 for models trained with different supervision strategies without finetuning. } 
\label{tab:generalizabilityISIC2018}
\footnotesize
\renewcommand{\arraystretch}{1.3}{
\begin{tabular}{@{}cccccc@{}}
\toprule
Methods & Data & Dice & Jaccard & Accuracy & Sensitivity \\ \midrule
\multirow{2}{1.5cm}{\centering UNet} & mask &86.78 &78.27 &92.69 & 93.85\\
 &BPAnno &86.68 &78.06 &92.92 & 94.64 \\
\hdashline 
\multirow{2}{1.5cm}{\centering UNet++} & mask  &87.11 &78.95 &92.63 & 94.64 \\
 &BPAnno  &86.67 &77.88 &92.7 & 95.12 \\
 \hdashline 
\multirow{2}{1.5cm}{\centering DeepLabV3+} & mask  &87.02 &78.73 &92.87 & 94.64 \\
 &BPAnno  &86.63 &77.98 &92.73 & 94.75 \\
 \hdashline 
\multirow{2}{1.5cm}{\centering TransUNet} & mask &86.34 &77.39 &92.38 &95.99 \\
&BPAnno  &86.43 &77.49 &92.55 & 95.93\\
\hdashline 
\multirow{2}{1.5cm}{\centering TransFuseS} & mask &87.67 &79.58 &93.22 & 95.04 \\
&BPAnno  &87.57 & 78.81 &94.92 & 93.39 \\
\hdashline 
\multirow{2}{1.5cm}{\centering HiFormer} & mask &87.27 &79.16 &93.10 &95.10 \\
&BPAnno  &87.44 &79.12 &92.87 &94.94 \\ \bottomrule
\end{tabular} }
\end{table}

\subsection{Error Analysis}  
\label{suc:4.4}
\begin{figure}[!h]
\centering
\includegraphics[width=0.45\textwidth]{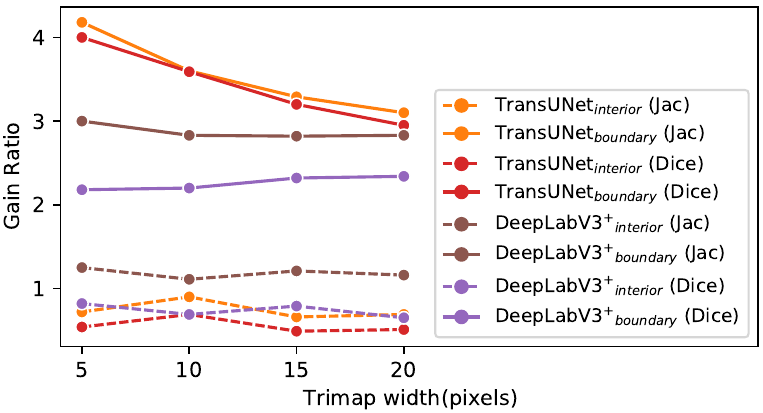}
\caption{Error analysis on the ISIC2017 test set. Both inside and outside a band of specific width are illustrated.}
\label{fig:trimaps}
\end{figure}

\begin{figure}[!h]
\centering
\begin{minipage}{0.2\textwidth}
\centering
\includegraphics[width=\linewidth]{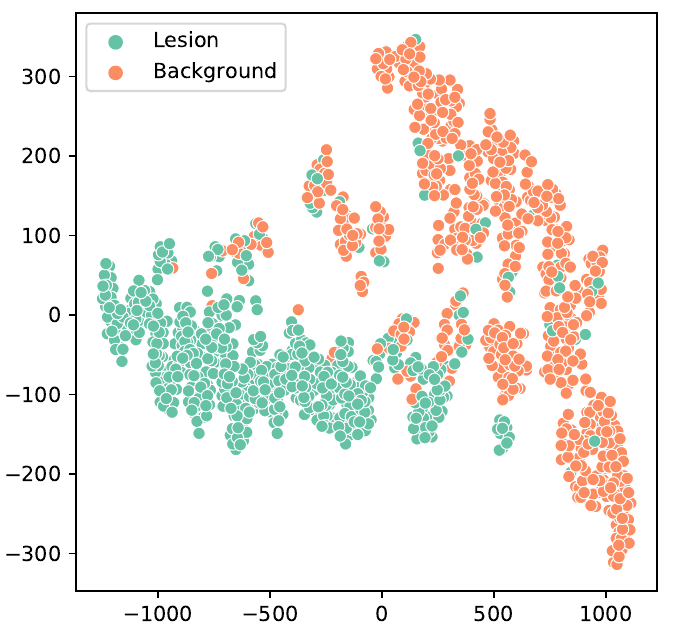}
\label{fig:subfig3sub1}\scriptsize{(a)}
\end{minipage}
\hspace{0.1cm}
\begin{minipage}{0.2\textwidth}
\centering
\includegraphics[width=\linewidth]{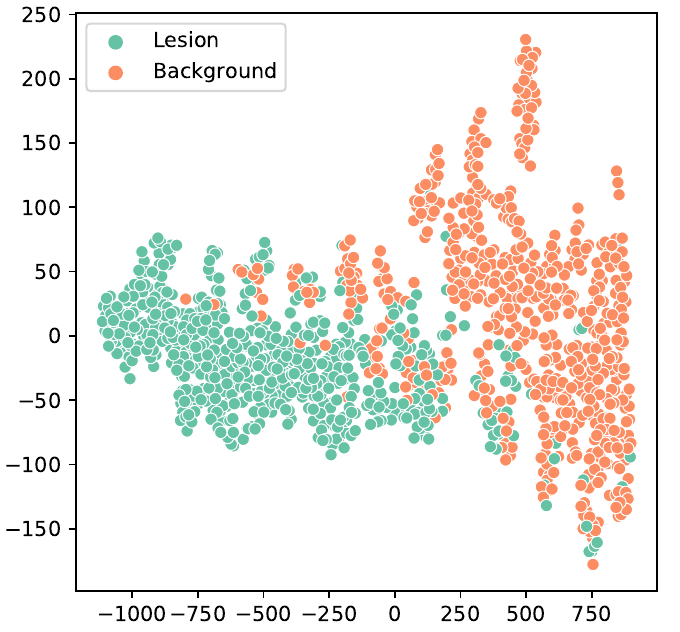}
\label{fig:subfig3sub2}\scriptsize{(b)}
\end{minipage}
\hspace{0.1cm}
\begin{minipage}{0.2\textwidth}
\centering
\includegraphics[width=\linewidth]{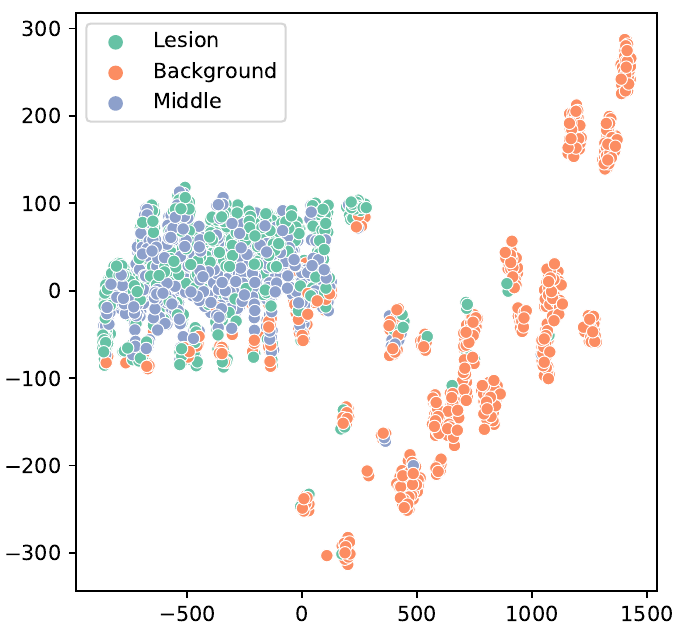}
\label{fig:subfig3sub3}\scriptsize{(c)}
\end{minipage}
\hspace{0.1cm}
\begin{minipage}{0.2\textwidth}
\centering
\includegraphics[width=\linewidth]{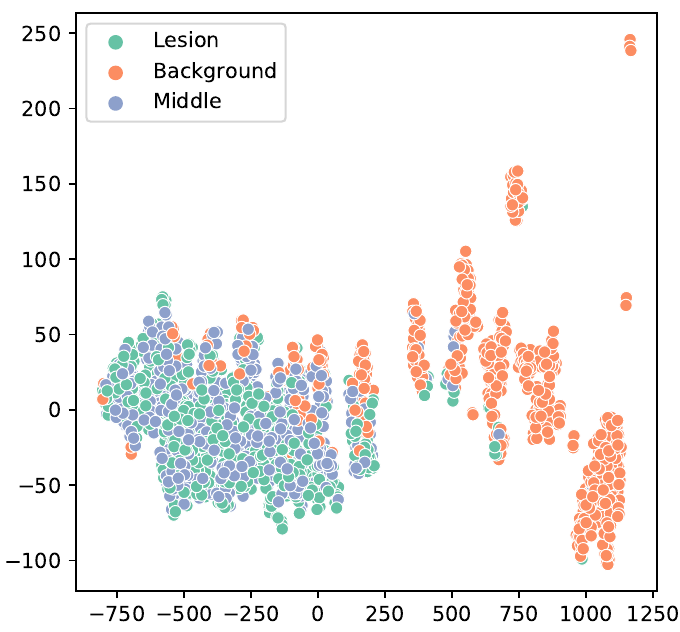}
\label{fig:subfig3sub4}\scriptsize{(d)}
\end{minipage}
\caption{t-SNE visualizations on ISIC2017 test set for the TransUNet and EAUWSeg(TransUNet). (a) and (b) display feature embedding generated by the constructed EAUWSeg and the TransUNet, respectively, with distinct colors representing the foreground and background. (c) and (d) illustrate the feature embedding generated by the constructed EAUWSeg(TransUNet) and TransUNet, with a separate focus on the lesion boundary.}
\label{fig:fig5}
\end{figure}

The proposed bounded polygon annotation has the advantage of explicitly providing prior emphasis on lesion boundaries during model training. To reveal this, following \cite{dai2015boxsup}, we separately evaluate the results in boundary and interior regions. 
Fig. \ref{fig:trimaps} illustrate the Jaccard and Dice score improvement achieved by our EAUWSeg compared to the BPAnno-supervised baselines, both inside and outside a band of specific width, referred to as boundary and interior regions. 
It can be seen that EAUWSeg consistently enhances the performance of the baseline models in both the boundary and interior regions, regardless of the trimap width. Specifically, our EAUWSeg achieves a substantial gain of over 2\% in performance within the boundary regions. This reveals the proposed EAUWSeg in capturing the intricate details of the boundary, which is attributed to our developed confidence-auxiliary consistency learner. Furthermore, we also illustrates the t-SNE visualization results of constructed feature space for TransUNet trained in fully-supervision setting and EAUWSeg in BPAnno-supervision manner. From Fig. \ref{fig:fig5}, it can be seen that our method can construct more compact feature space compared with the fully-supervised baseline, especially in the lesion boundary.  

\subsection{Annotation Cost Analysis}
To reveal the annotation cost decreasing ability of the proposed bounded polygon annotations, we conduct the comparison study focusing on the annotation workload. In this study, a dermatologist with over ten years of experience from a general hospital in the central city performs two types of annotations, i.e., pixel-to-pixel dense annotation and the proposed bounded polygon annotation, on twenty image selected in ISIC2017. It takes an average of 55 and 10 seconds for the pixel-to-pixel dense annotation and bounded polygon annotation, respectively. This indicates that annotating bounded polygon of skin lesion in dermoscopic image requires only 18\% of annotation cost when compared with the pixel-to-pixel annotations. Combining the experimental results illustrated in Table \ref{tab:totalresults} and Table \ref{tab:generalizabilityISIC2018}, the proposed method delivers superior performance and comparable generalization ability when compared to its fully-supervised counterparts. These results reveal that bounded polygon annotations coupled with EAUWSeg can be a cost-effective solution for weakly-supervised medical image segmentation.



\section{Conclusion and Future Work}
\label{sec:sec5}
In this work, to eliminate the annotation uncertainty existed in weakly-supervised medical image segmentation, we propose the bounded polygon annotation, in which label only two polygons while providing promising prior of lesion boundary during training. To further eliminate the uncertainty included in the bounded polygon as well as to leverage the prior emphasis delineated by bounded polygons, we develop EAUWSeg, a learning framework tailored for bounded polygon that include a confidence-auxiliary consistency incorporated with a classification-guided confidence generator is designed to provide reliable supervision signal for pixels in uncertain region. Extensive experimental results demonstrate that EAUWSeg can not only outperform existing weakly-supervised segmentation methods but also delivers superior performance compared to fully-supervised counterparts, with less than 20\% of the annotation workload. 

This work is a preliminary attempt to focus on eliminating annotation uncertainty in weakly-supervised medical image segmentation. Extensive experimental results have demonstrated its cost-efficient and effectiveness of the bounded annotation, while there is still several limitations. This study mainly focuses on the weakly-supervised medical image segmentation in binary case. When applied to the instance segmentation, it may suffer from some challenges, such as encompassing pixels belong to foreground with different categories in the envelope-like polygon. In future work, we will focus on solving this kind of problems. 

\bibliographystyle{IEEEtran}
\bibliography{IEEEabrv,ref}

\end{document}